\documentclass[conference]{IEEEtran}
\IEEEoverridecommandlockouts
\usepackage{cite}
\usepackage{amsmath,amssymb,amsfonts}
\usepackage{algorithmic}
\usepackage{booktabs}
\usepackage{bm}
\usepackage{makecell}
\usepackage{graphicx}
\usepackage{textcomp}
\usepackage{xcolor}
\usepackage[caption=false,font=normalsize,labelfont=sf,textfont=sf]{subfig}
\usepackage[hidelinks]{hyperref}
\def\BibTeX{{\rm B\kern-.05em{\sc i\kern-.025em b}\kern-.08em
    T\kern-.1667em\lower.7ex\hbox{E}\kern-.125emX}}
\usepackage{orcidlink}
\begin{document}

\title{Safe Local Navigation for Ackermann-Steered Robots in Unmapped Environments\\

\thanks{The authors are with the Department of Electrical and Computer Engineering, McMaster University, Hamilton, ON L8S 4L8, Canada (e-mail: schaiblc@mcmaster.ca; sirous@mcmaster.ca).}

\thanks{\textbf{Code}: \href{https://github.com/schaiblc/McMaster_AEV_MPC_Algorithms}{https://github.com/schaiblc/McMaster\_AEV\_MPC\_Algorithms}}
}

\author{\IEEEauthorblockN{Christian Schaible \orcidlink{0009-0003-8811-4335}, Shahin Sirouspour \orcidlink{0000-0003-4882-2161}}}

\maketitle

\begin{abstract}
A control framework is proposed for safe local navigation of mobile robots equipped with Ackermann steering in unmapped environments where a global goal is absent. Based on local obstacle detections, the safest heading angle is determined along the direction of the largest open space ahead of the vehicle. Guided by this direction, bounding lines are constructed on the left and right sides of the vehicle to achieve obstacle separation. These bounding lines are obtained by solving a convex quadratic optimization that maximizes vehicle-to-obstacle clearance. Optionally, conditions are imposed on the bounding lines to preserve parallelism and smooth abrupt changes from prior control steps. A feedback-linearizing controller is then used to regulate the vehicle's distance from one or both bounding lines, effectively enabling tracking of a local reference path that preserves safety through obstacle clearance maximization. Open-source code is included for the application of this control scheme. Experimental results demonstrate that the proposed method produces safer navigation paths with significantly shorter computation times, compared to some existing exploration-based planners.
\end{abstract}

\section{Introduction}
Autonomous navigation in unstructured environments requires a holistic framework for real-time decision making that extends to arbitrary settings and applications. However, path planning in autonomous vehicle and mobile robot contexts often incorporates knowledge of a mapped environment or a known global goal position \cite{Jafarzadeh18,Dong2024}. Certain applications, such as search and rescue, exploration or particular racing situations, require safe, effective local navigation with no prior or global information. In these cases, the objective changes from tracking a global reference path to local planning in the most sensible direction of the largest open space while preserving safety through a high distance to nearby obstacles. This presents a universal foundation for safe local navigation, which can be incorporated in hierarchical planning to achieve full global navigation.

For real-time planning to be possible, computation times must be within the control rate at all times, even in the worst case. Faster control rates enable quicker reactions to be made when sudden changes in the environment occur, such that safety and collision-free navigation can be upheld. Improvements in modern processing and computational power have motivated the development of more sophisticated, higher-dimensional predictive planners such as those based on Model Predictive Control (MPC) \cite{Ji2017,MPC_Planner}. For these planners, exploiting insights into the problem is vital for effective solutions to be found in real time \cite{Lam2010}. For vehicles approximated by the kinematic bicycle model \cite{Rajamani2012}, nonholonomic vehicle constraints cause path planning optimizations to be non-convex and non-linear \cite{MPC_Planner,Williams2016}. This can be addressed through methods such as linearizations \cite{Alcal2019}, which risk compromising accuracy, or optimization stopping criteria to maintain real-time control. Alternatively, simpler non-predictive methods present an opportunity to guarantee fast navigation solutions; however, degradations in performance arise \cite{Anil23} and must be minimized.

\begin{figure}[t]
  \centering
  \includegraphics[width=\linewidth]{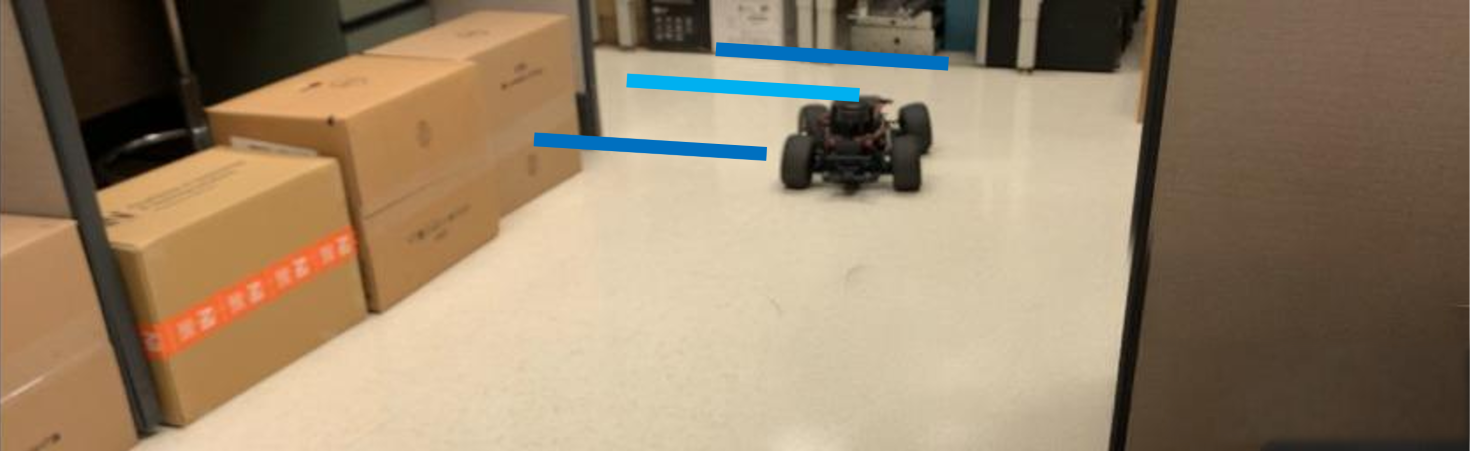}\\
  \includegraphics[width=\linewidth]{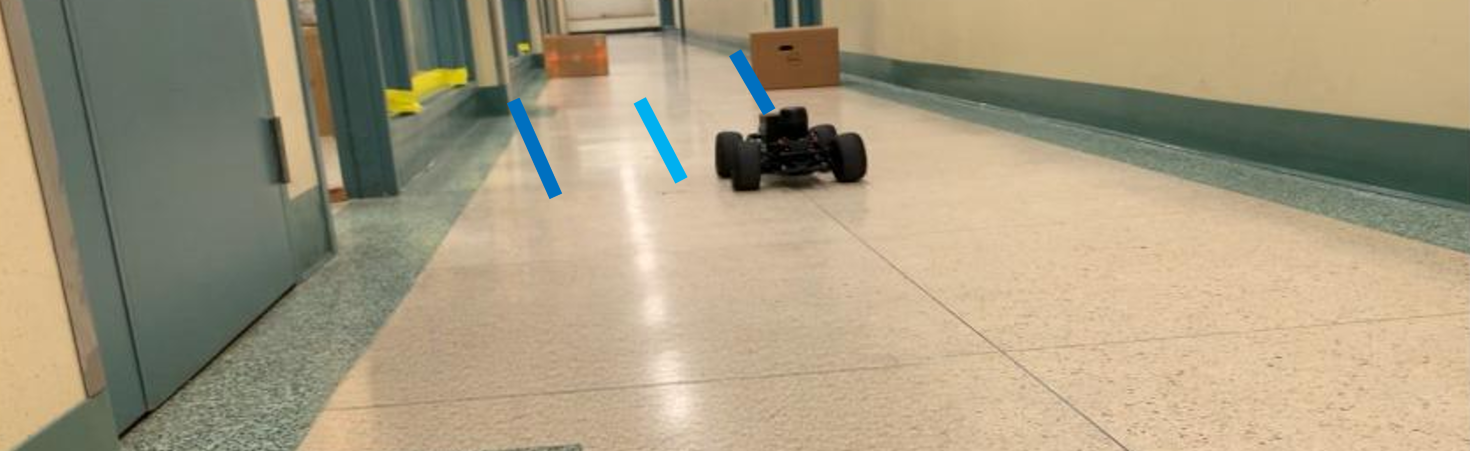}\\
  \includegraphics[width=\linewidth]{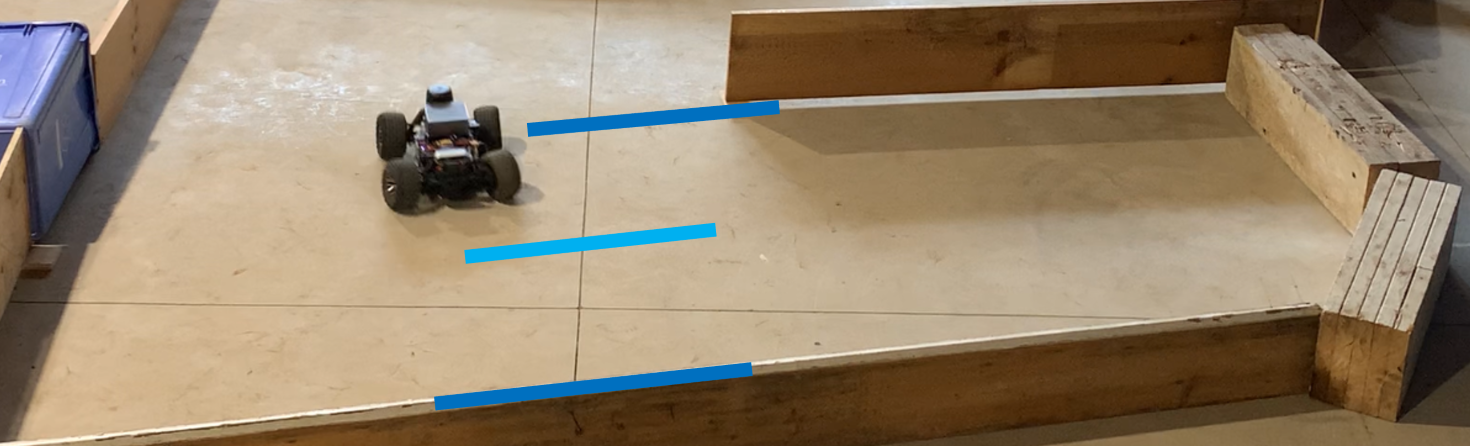}
  \caption{Local linear reference path (light blue) and obstacle bounding lines (dark blue) at three distinct control steps across varying environments.}
  \label{PD_lines}
\end{figure}

One low-cost solution is the use of Proportional-Integral-Derivative (PID) \cite{Emirler2014} or Proportional-Derivative (PD) steering control. Here, navigation performance is subject to gain tuning \cite{Mourad22} and the selection of an optimal reference path. This paper presents a framework for generating optimal local obstacle bounding lines and thereby reference paths in real-time \mbox{(Fig. \ref{PD_lines})} before employing a feedback-linearizing PD controller to ensure vehicle actuation along the reference path.

The main contribution of this approach is the ability to achieve standalone local navigation in unknown environments with low planning times while promoting safety through obstacle clearance maximization. The proposed method outperforms existing exploration-based planners across safety, control effort and computation time metrics in real-world experiments, illustrating its value for navigation in unknown environments. Evaluation is conducted both in simulation and experimentally using a 1/10$^{\text{th}}$ scale RC vehicle, while the C++/ROS code designed for implementation of the control scheme is released as open-source.

\section{Related Work}
Given a reference path to track, PID control presents a simple yet effective way to achieve reactive navigation according to vehicle kinematics. Using lateral and longitudinal dynamic vehicle models, one study designs a robust PID steering controller, which attains path following of simple, simulated curves \cite{Emirler2014}. Double-axle steering is achieved in a separate path-tracking approach, where for linear or circular arc paths and a constant velocity, a linear, time-invariant, decoupled PID controller is employed \cite{DeSantis95}. Another approach constructs the path from an initial location to a goal position through piecewise lines between waypoints before approaching path following through transverse feedback linearization. Performance of this approach in a simple experiment for a differential drive mobile robot is shown to be comparable to an MPC strategy \cite{DSouza21}. Optimal tuning for PID steering control has been explored in heuristic and tracking error-based approaches to achieve improved path following at different speeds \cite{Farag18}.

An alternative path tracking method is through pure pursuit \cite{Wallace-1985}. This approach uses a chosen look-ahead distance to determine the necessary curvature that will move the vehicle to the identified point on the reference path. Similar to PID control, computation time and complexity are low. Meanwhile, extensions have attempted to improve tracking error and adaptive performance through varying look-ahead distances \cite{Huang2020} and linear velocities \cite{Macenski23}.

The creation of a reference path is typically done by a global planner, which connects the start and goal points \cite{LaValle,Kavraki96}, before a local planner revises a local path which tracks the global path while performing obstacle avoidance \cite{Chikurtev2020,Liang25}. In the case of unknown environments, however, a framework for path generation in the absence of a goal is required. One method introduces a diffusion policy incorporating a high-capacity visual transformer encoder to achieve goal-agnostic exploration and navigation towards visually indicated goals in novel settings \cite{Sridhar24}. Alternatively, the Artificial Potential Field (APF) method determines the best local path for obstacle avoidance through an expression of attractive and repulsive forces \cite{Khatib}. Attractive forces are contingent on a known goal location, though, and APF planning is highly susceptible to falling into local minima solutions where passage failure can arise \cite{Koren}.

Predictive navigation increases planning flexibility at the cost of increased complexity. MPC is prominently used for motion control and path planning, where one open-source MPC local planner \cite{MPC_Planner} can perform collision avoidance while reaching intermediate goal positions along a globally known path. Another method generates front steering angle commands by tracking a potential-field-based reference trajectory with multiconstrained MPC \cite{Ji2017}. The Dynamic Window Approach (DWA) \cite{DWA} offers local planning by parameterizing the local path with constant translational and rotational velocity pairs. Vehicle dynamics limit the feasible velocity pairs before each is evaluated in an optimization that prioritizes a heading towards the goal location, clearance of nearby obstacles and a larger forward velocity using fixed cost weights. 

Another local planning method uses an elastic band algorithm, which deforms the global path to avoid obstacles while maintaining its overall shape \cite{EB}. The Timed Elastic Band (TEB) planner \cite{TEB} extends this by incorporating vehicle dynamics constraints, where local solutions are found through large-scale optimizers for sparse systems. Shorter paths are encouraged, risking closer obstacle proximity and higher collision likelihood, especially in uncertain, dynamic conditions.



\section{Proposed Local Navigation Method}\label{sec3}
This section presents a novel control framework for local navigation in unknown environments. The construction of local obstacle bounding lines and thereby a local reference path is proposed through a Quadratic Program (QP) for efficient, real-time optimization and obstacle clearance maximization. Path following is then proposed using a PD steering controller with feedback linearization, based on the current speed.

\subsection{Bounding Lines and Reference Path Formulation}
The unknown environment is constructed in the moving vehicle base frame through a set of detected obstacles (in practice, through a sensor such as LiDAR). The proposed coordinate system assigns $+x$ as the forward vehicle direction, $+y$ as the leftward vehicle direction and $+\theta$ as the counterclockwise rotation direction, beginning from the $+x$ axis. To determine the local reference path, a heading angle is first found, inspired by the Follow the Gap Method (FGM) \cite{Sezer2012}. Here, obstacles $\mathcal{O}=\{{(x_{i}, y_{i})}\}_{i=0}^{N_{{\text{obs}}}-1}$ are sorted in order of increasing orientation $\theta_{i}$ with respect to the ego vehicle before the largest range $r_{i}$-weighted angular gap containing only obstacles further than a threshold $d_{\textup{safe}}$ is found:

\begin{equation}
(i^*\mkern-3mu,\mkern-2muk^*\mkern-1mu)\!\mkern-2mu=\!\mkern-1mu
\arg\mkern-5mu\max_{(i, k) :\ \mkern-5mu \theta_j \in [\mkern-1mu-\mkern-1mu\frac{\pi}{2}\mkern-1mu,\mkern-1mu\frac{\pi}{2}\mkern-1mu],\ \mkern-6mu r_j > d_{\mathrm{safe}}\ \mkern-6.5mu \forall j\in \{i,...,k\}}\mkern-4mu
\sum_{j=i}^{k} \!\frac{\mkern-1mur_j\mkern-1mu (\mkern-1mu\theta_{j\mkern-1mu+\mkern-1mu1} \!\mkern-3mu-\!\mkern-1mu \theta_{j\mkern-1mu-\mkern-1mu1}\mkern-2mu)}{2}
\end{equation}
where only obstacles in the front $\pi$ radian window are considered. Using the safest angular gap's start $\theta_{\textup{start}}=\theta_{i^*}$ and end $\theta_{\textup{end}}=\theta_{k^*}$ angles, the heading direction $\theta_{\textup{head}}$ becomes:
\begin{equation}
    \theta_{\text{head}}=\frac{\theta_{\text{start}}+\theta_{\text{end}}}{2}
\end{equation}

Now, the obstacle set is divided into left, right and excluded clusters with respect to the heading angle $\theta_{\text{head}}$. The $N_{\textup{left}}$ obstacles that satisfy \mbox{$\theta_{i} \mathbin{\in} [\theta_{\text{head}}+a_l,\theta_{\text{head}}+b_l]$} compose the left cluster $\mathcal{O}_l$ while the right cluster $\mathcal{O}_r$ consists of the $N_{\textup{right}}$ obstacles where \mbox{$\theta_{i} \mathbin{\in} [\theta_{\text{head}}-b_r,\theta_{\text{head}}-a_r]$} holds. The QP for the local obstacle bounding lines is formulated in a manner analogous to Support Vector Machines (SVMs) \cite{Cortes1995}. Here, decision boundaries are identified that best separate classes of points by maximizing the margin between them. This method has been applied in global planning to maximize distance to obstacles along paths between start and goal points \cite{Miura2006,Morales2016}.

In general, an obstacle bounding line can be described as $w^Tp+b=0$, where normalization produces $w^Tp+1=0$ since the boundary will not pass through the vehicle base frame origin ($b\neq0$). Here, $w=[w_x,w_y]^T$ and the obstacle point $p\!=\![x,y]^T\!\!\in\! \mathcal{O}$. The distance $d$ from $p$ to the bounding line is:
\begin{equation}
    d=\frac{| w^Tp+1 |}{\sqrt{w^Tw}}
\end{equation}
where taking $p=[0,0]^T$ as the base frame origin, the vehicle-to-line distance $d_0$ becomes:
\begin{equation}
    d_0=\frac{1}{\sqrt{w^Tw}}
\end{equation}
Therefore, the QP that maximizes distance from the origin to the bounding line while ensuring all obstacle points remain on the separated side can be expressed as:
\begin{equation}
\begin{aligned}
\min_{w} \quad & \frac{1}{2}w^Tw \\
\text{subject to} \quad & w^Tp_i+1 \leq 0 \quad \forall p_i \in \mathcal{\tilde O}
\end{aligned}
\end{equation}
which holds for both $\mathcal{\tilde O}=\mathcal{O}_l$ and $\mathcal{\tilde O}=\mathcal{O}_r$ to construct the unique left $w_l$ and right $w_r$ bounding lines, respectively.

To ensure smoothness and prevent sudden changes in bounding lines from the prior control step, an augmented objective for the QP is introduced:
\begin{equation}
    \min_{w_k} \quad \frac{1}{2}\alpha w_k^Tw_k+\frac{1}{2}(1-\alpha)(w_k-w_{k-1})^T(w_k-w_{k-1})
\end{equation}
where $k$ denotes the current time step, $w_k$ represents the current bounding line parameters and $w_{k-1}$ is the known bounding line from the prior time step. The weight $\alpha$ balances between maximizing the current obstacle bounding line clearance and minimizing the difference in successive bounding lines. This objective can be simplified, producing the new QP for smoothed changes in bounding lines:
\begin{equation}
\begin{aligned}
\min_{w_k} \quad & \frac{1}{2} w_k^Tw_k+(\alpha-1)w_{k-1}^Tw_k \\
\text{subject to} \quad & w_k^Tp_i+1 \leq 0 \quad \forall p_i \in \mathcal{\tilde O}\label{smoothQP}
\end{aligned}
\end{equation}

Considering the unconstrained problem, the closed-form solution can be found by setting the gradient of the objective function in \eqref{smoothQP} to zero:
\begin{align}
    &\frac{\partial}{\partial w_k}(\frac{1}{2} w_k^Tw_k+(\alpha-1)w_{k-1}^Tw_k)=w_k+(\alpha-1)w_{k-1}=0
\end{align}
Through recursive expansion, the current bounding line can be described using the initial bounding line parameters $w_0$ as the response of a first-order discrete-time system through $w_k=(1-\alpha)^kw_0$. Equivalently, the continuous-time system is represented by $w(t)=e^{-\frac{t}{\tau}}w_0$. By setting $t=k\Delta t$, where $\Delta t$ is the discrete timestep, the discrete-time parameter can be related to the continuous-time time constant $\tau$ via \mbox{$\alpha=1-e^{-\frac{\Delta t}{\tau}}$}. This formulation allows the smoothness of successive obstacle bounding lines to be adjusted by directly selecting the time constant $\tau$, thereby balancing smooth transitions against obstacle clearance maximization within the optimization.

\begin{figure}[ht]
    \centering
    \includegraphics[width=0.95\columnwidth]{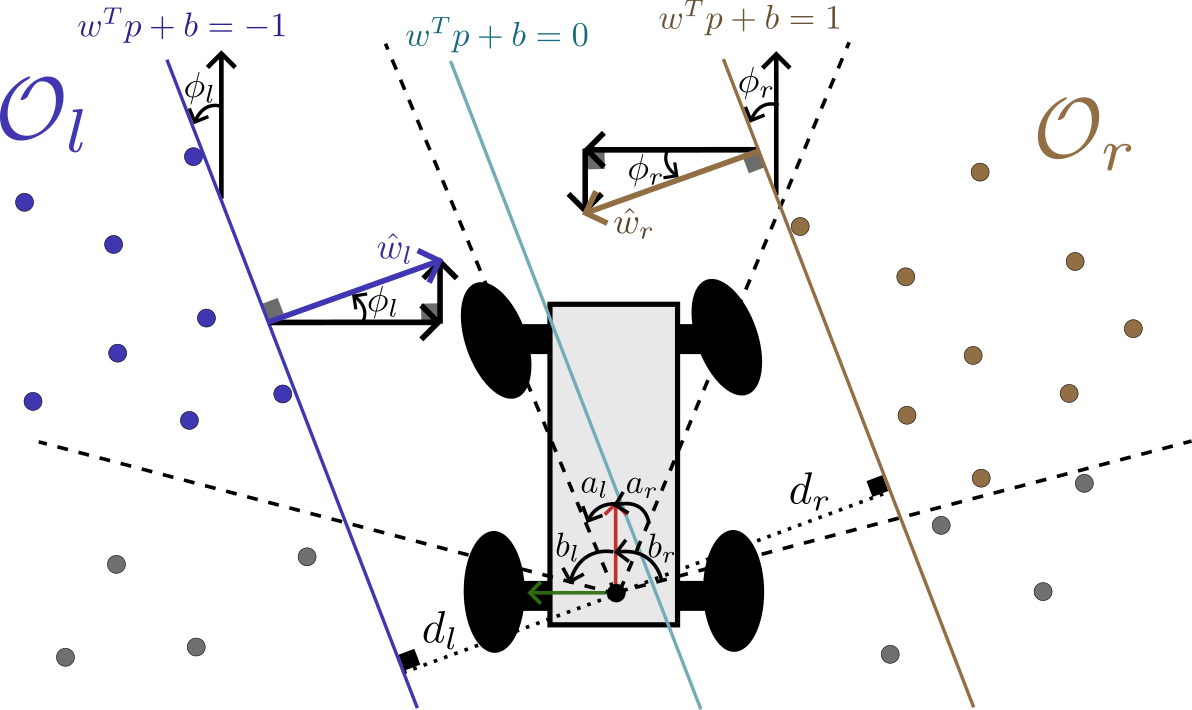}
    \caption{Left and right parallel bounding lines, obstacle clusters and the optimal center line that maximizes obstacle clearance. The case of independent bounding lines is described similarly.}
    \label{QP_boundlines}
\end{figure}

Alternatively, the bounding lines can be constrained to be parallel, as in the standard SVM formulation, to define a consistent maximum margin and preserve symmetry. Now, the left $w^Tp+b\!=\!-1$ and right $w^Tp+b\!=\!1$ \mbox{bounding lines} (Fig. \ref{QP_boundlines}) are characterized by the shared parameters $w\!\in\! \mathbb{R}^2, b\!\in\!\mathbb{R}$. The distance between the parallel lines becomes $\frac{2}{\sqrt{w^Tw}}$, while the shared line parameters mean the obstacle separating inequalities must respect direction in the ensuing optimization:
\begin{equation}
\begin{aligned}
\min_{w,b} \quad & \frac{1}{2}w^Tw \\
\text{subject to} \quad & w^Tp_i+b-1 \geq 0 \quad \forall p_i \in \mathcal{O}_r\\
                        & w^Tp_j+b+1 \leq 0 \quad \forall p_j \in \mathcal{O}_l\\
                        & -1+\epsilon \leq b \leq 1-\epsilon\\
\end{aligned}
\label{quadprogopt}
\end{equation}

Here, $b$ is bounded using a small, positive $\epsilon$ to ensure that the vehicle origin remains between the two obstacle bounding lines. In practice, strict convexity is ensured by adding a small quadratic objective term involving $b$. The center line is defined by $w^Tq+b=0$, serving as the effective tracking line, where normalization by $b$ leads to only two line parameters. Goldfarb and Idnani's active-set dual method \cite{Goldfarb1983} can be used to solve the QP with linear inequality constraints, producing the bounding and center lines for ensuing PD control.

\subsection{Feedback Linearization-Based Vehicle Control}
Now, PD control regulates the vehicle's distance from each bounding line, where feedback linearization establishes closed-loop system dynamics corresponding to those of a standard second-order linear time-invariant system. Using independent bounding lines, $w_l$ and $w_r$ are generated separately as described, whereas in the parallel bounding line case, variable reduction similarly produces $w_l=\frac{w}{b+1}$ and $w_r=\frac{w}{b-1}$. Respectively, the left and right vehicle-to-line distances become $d_l=\frac{1}{\sqrt{w_l^Tw_l}}$ and $d_r=\frac{1}{\sqrt{w_r^Tw_r}}$. Each normalized vector is subsequently represented by $\hat w_l=d_lw_l$ and $\hat w_r=d_rw_r$.

In order for the left bounding line to remain on the vehicle's left, $w_{y,l}<0$, while similarly for the right bounding line to remain on the right, $w_{y,r}>0$. Therefore, each line's unit normal points inwards, towards the vehicle. Using the coordinate convention, the unit normal vectors are projected onto each axis through:
\begin{align}
    \sin \phi_l&=\begin{bmatrix}1&0\end{bmatrix}\cdot\hat w_l, &\sin \phi_r&=\begin{bmatrix}-1&0\end{bmatrix}\cdot \hat w_r\\
    \cos \phi_l&=\begin{bmatrix}0&-1\end{bmatrix}\cdot\hat w_l, &\cos \phi_r&=\begin{bmatrix}0&1\end{bmatrix}\cdot \hat w_r
\end{align}
where $\phi_l$ and $\phi_r$ respectively describe the left and right bounding line orientations with respect to the $+x$ axis. The rates of change for the left and right vehicle-to-line distances can now be found by projecting the current vehicle speed $v$ in the $+x$ direction back onto the bounding line normal vectors:
\begin{equation}
    \dot d_l=v\sin\phi_l,\quad \dot d_r=-v\sin\phi_r\label{first_order}
\end{equation}
Taking the time derivative of each, the second-order rates of change are:
\begin{equation}
    \ddot{d}_l=-\frac{v^2}{l}\cos \phi_l\tan \delta,\quad \ddot{d}_r=\frac{v^2}{l}\cos \phi_r\tan \delta\label{secondorder}
\end{equation}
Here, the current bounding line orientation derivatives are subject to the moving vehicle base frame's orientation derivative $\dot \theta$ through $\dot \phi_l=\dot \phi_r=-\dot \theta$. According to the kinematic bicycle model \cite{Rajamani2012}, $\dot \theta=\frac{v}{l}\tan(\delta)$ with fixed wheelbase $l$ and current steering angle $\delta$.

These second-order non-linear dynamics can be employed to either control the desired distance $d_l^\textup{des}$ to the left bounding line using tracking error $\tilde d_l=d_l-d_l^\textup{des}$ or $d_r^\textup{des}$ to the right bounding line with tracking error $\tilde d_r=d_r-d_r^\textup{des}$. Alternatively, the vehicle can be controlled to remain midway between the bounding lines using $d_{lr}=d_l-d_r$. Using \eqref{first_order}, $\dot d_{lr}=\dot d_l-\dot d_r$ while through \eqref{secondorder}, $\ddot{d}_{lr}=\ddot{d}_l-\ddot{d}_r$. The second-order linear time-invariant system is constructed using a desired distance $d_{lr}^\textup{des}$ where $d_{lr}^\textup{des}=0$ ensures maximum, equal clearance from both bounding lines:
\begin{equation}
    k_pd_{lr}^\textup{des}=\ddot{d}_{lr}+k_d\dot d_{lr}+k_pd_{lr}
\end{equation}
The proportional $k_p$ and derivative $k_d$ gains influence the system response, while the linear system is obtained from the non-linear dynamics through feedback linearization. Here, the linearizing steering angle control input is derived as:
\begin{equation}
    \delta_{lr}=\mathrm{atan}(\frac{l}{v^2(\cos\phi_l+\cos\phi_r)}(k_d\dot d_{lr}+k_p\tilde d_{lr}))\label{PD_feedback}
\end{equation}
with tracking error $\tilde d_{lr}=d_{lr}-d_{lr}^\textup{des}$ and $v>0$ to avoid the singular case at $v=0$. The independent left and right bounding line control problems take similar forms, where linearization produces $\delta_l$ and $\delta_r$ for vehicle actuation, respectively:
\begin{align}
    \delta_{l}&=\mathrm{atan}(\frac{l}{v^2\cos\phi_l}(k_d\dot d_{l}+k_p\tilde d_{l}))\label{PD_feedback_l}\\
    \delta_{r}&=\mathrm{atan}(\frac{-l}{v^2\cos\phi_r}(k_d\dot d_{r}+k_p\tilde d_{r}))\label{PD_feedback_r}
\end{align}

The forward speed $v$ of the vehicle is set to a nominal value $v_0$, where in the case of high collision-risk obstacles directly ahead, the minimum distance $d_\textup{min}$ enforces vehicle slowdown:
\begin{equation}
    v=v_0(1-e^{-\frac{\max(d_\textup{min}-d_\textup{stop},0)}{\alpha_v}})
\end{equation}
with 
\begin{equation}
d_{\min} = \min_{\substack{i\in\{0,...,N_{\textup{obs}-1}\} \\ -\theta_{\textup{FOV}} \leq \theta_i \leq \theta_{\textup{FOV}}}} r_i
\end{equation}
Only obstacles within the front field of vision defined by $\theta_{\textup{FOV}}$ are considered, where $d_\textup{stop}$ denotes the obstacle distance at which the vehicle is forced to stop. The decay rate $\alpha_v$ controls how sharp the speed decrease is as $d_{\textup{min}}$ is reduced. Finally, the $\delta_\textup{cmd}$ and $v_\textup{cmd}$ control inputs are limited to satisfy feasible vehicle dynamics according to magnitude ($\pm\delta_\textup{max},v_\textup{max}$) and rate of change ($\pm\Delta\delta_\textup{max},\pm\Delta v_\textup{max}$) limits.

Using this PD control approach, the vehicle steering angle $\delta_\textup{cmd}$ and forward velocity $v_\textup{cmd}$ control inputs are thus generated and applied for vehicle actuation. The process of finding a heading angle, bounding lines and control inputs through feedback linearization and minimum forward obstacle distance is conducted again at the next control step $\Delta t$ for real-time local navigation.

\section{Simulation and Experimental Results}
\subsection{Evaluation Framework}
\begin{figure}[ht]
    \centering
    \includegraphics[width=0.98\columnwidth]{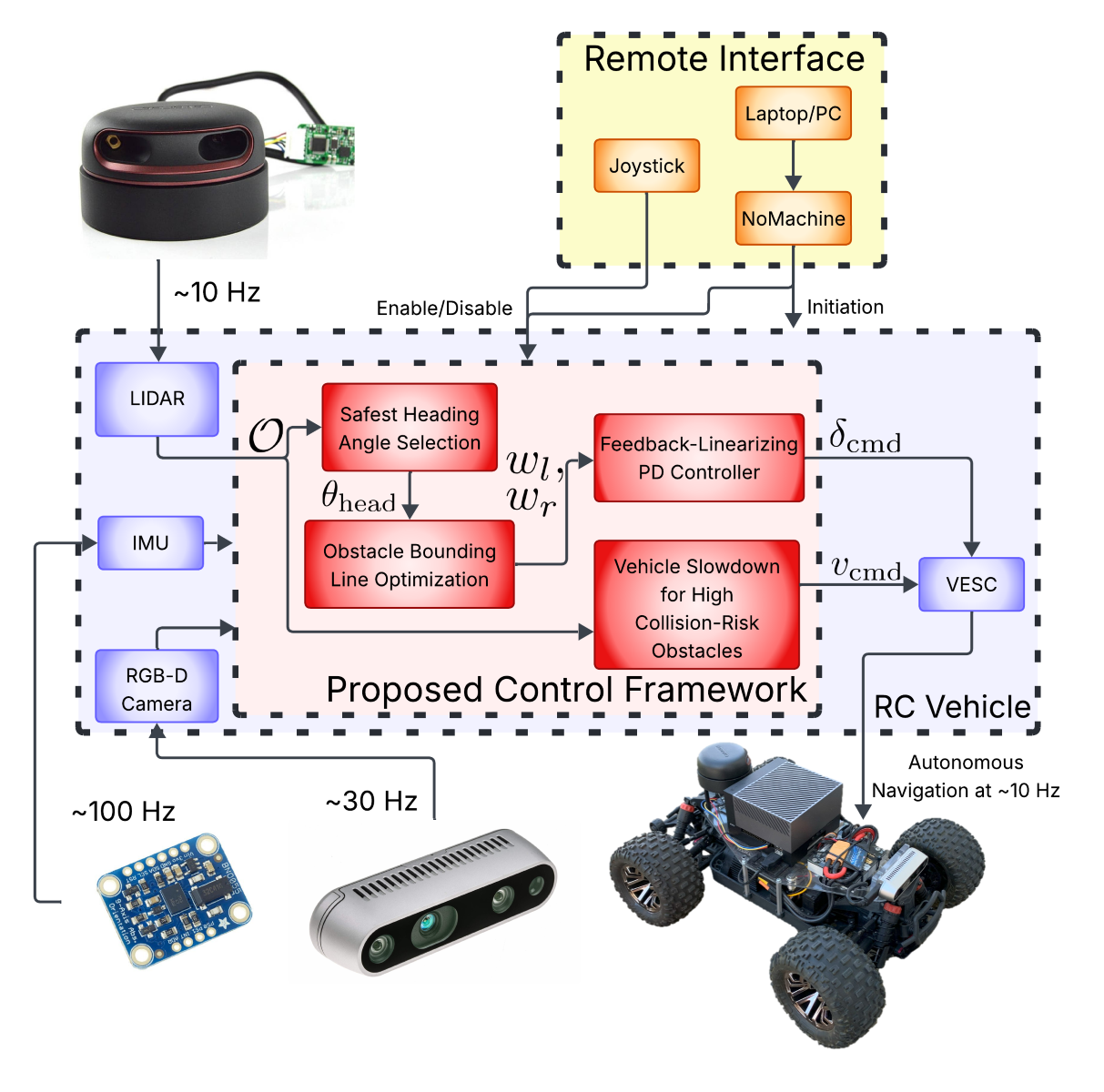}
    \caption{System integration is represented by a block diagram highlighting the sensors, navigation framework, actuation and remote interface with the target platform. In simulation, the sensor and actuation components are emulated.}
	\label{PD_blockdiagram}
\end{figure}

To validate performance, the proposed navigation framework is evaluated against existing local planners. Testing is conducted both in the \textit{f1tenth\_simulator} simulation environment \cite{Babu2020} and in real-world conditions on a $1/10^\textup{th}$ scale RC vehicle platform motivated by the F1TENTH testbed \cite{f1tenth}. The system hardware consists of the NVIDIA Jetson AGX Orin, RPLIDAR A2M8 laser scanner, BNO055 IMU and RealSense D435i \mbox{RGB-D} camera. Fig. \ref{PD_blockdiagram} depicts the system architecture through a block diagram with interconnected sensing, control, remote interfacing and actuation stages.

Here, the LiDAR sensor is primarily used for local obstacle detection, whereas the IMU and RGB-D camera contribute to more comprehensive perception through localization, ground-plane filtering, and obstacle identification. Remote interfacing is used to both initiate operation and enable autonomous navigation. The laser scanner, RGB-D camera and IMU module publish data at approximately 10 Hz, 30 Hz and 100 Hz, respectively. Ensuing vehicle actuation based on the control input commands $\delta_\textup{cmd}$ and $v_\textup{cmd}$ is provided by the electronic speed controller, VESC, at about 10 Hz. The open-source software stack that uses the proposed navigation approach is implemented in C++ using ROS Noetic.

The baseline local planning approaches adhere to the standard ROS navigation stack, in which the central \textit{move\_base} node connects a local and global planner via \textit{nav\_core} to perform navigation and obstacle avoidance using 2D occupancy grid costmaps. Since this paper focuses on navigation in unmapped environments, no predefined global goal is specified; instead, these planners utilize a common exploration package, \textit{explore\_lite}, which continuously updates unexplored frontiers. In these tests, the largest frontier located ahead of the vehicle is selected as the immediate goal.

The global planner then generates a path toward the time-varying goal, while the local planner handles collision avoidance, producing an optimized trajectory for immediate motion. In these tests, three common ROS local planners---\textit{dwa\_local\_planner} \cite{DWA}, \textit{teb\_local\_planner} \cite{TEB} and \textit{mpc\_local\_planner} \cite{MPC_Planner}---are evaluated with default settings, except for those corresponding to the kinematics and physical parameters of the specific experimental vehicle.

\subsection{Simulation Results}\label{simsec}
In the simulation environment, Section \ref{sec3}'s navigation approach is evaluated with pertinent parameters \mbox{$l=\textup{0.287 m}$}, $\delta_\textup{max}=\textup{0.4189 rad}$, $v_0=\textup{1.5 m/s}$, \mbox{$d_\textup{stop}=\textup{0.8 m}$}, $\alpha_v=\textup{0.5 m}$, $\theta_\textup{FOV}=\frac{\pi}{8}\, \textup{rad}$, \mbox{$k_p=3.5\, \textup{s}^{-2}$} and $k_d=4\, \textup{s}^{-1}$. For heading angle selection, \mbox{$d_\textup{safe}$ = 2 m}, $a_l\!\!=\!\!a_r\!\!=\!\!\frac{\pi}{9}\ \textup{rad}$ and $b_l\!\!=\!\!b_r\!\!=\!\!\frac{\pi}{2}\ \textup{rad}$. The parallel bounding line optimization in \eqref{quadprogopt} is used along with the dual bounding line control problem, culminating in the linearizing control input described by \eqref{PD_feedback}. 

\begin{figure}[ht]
    \centering
    \includegraphics[width=0.81\columnwidth]{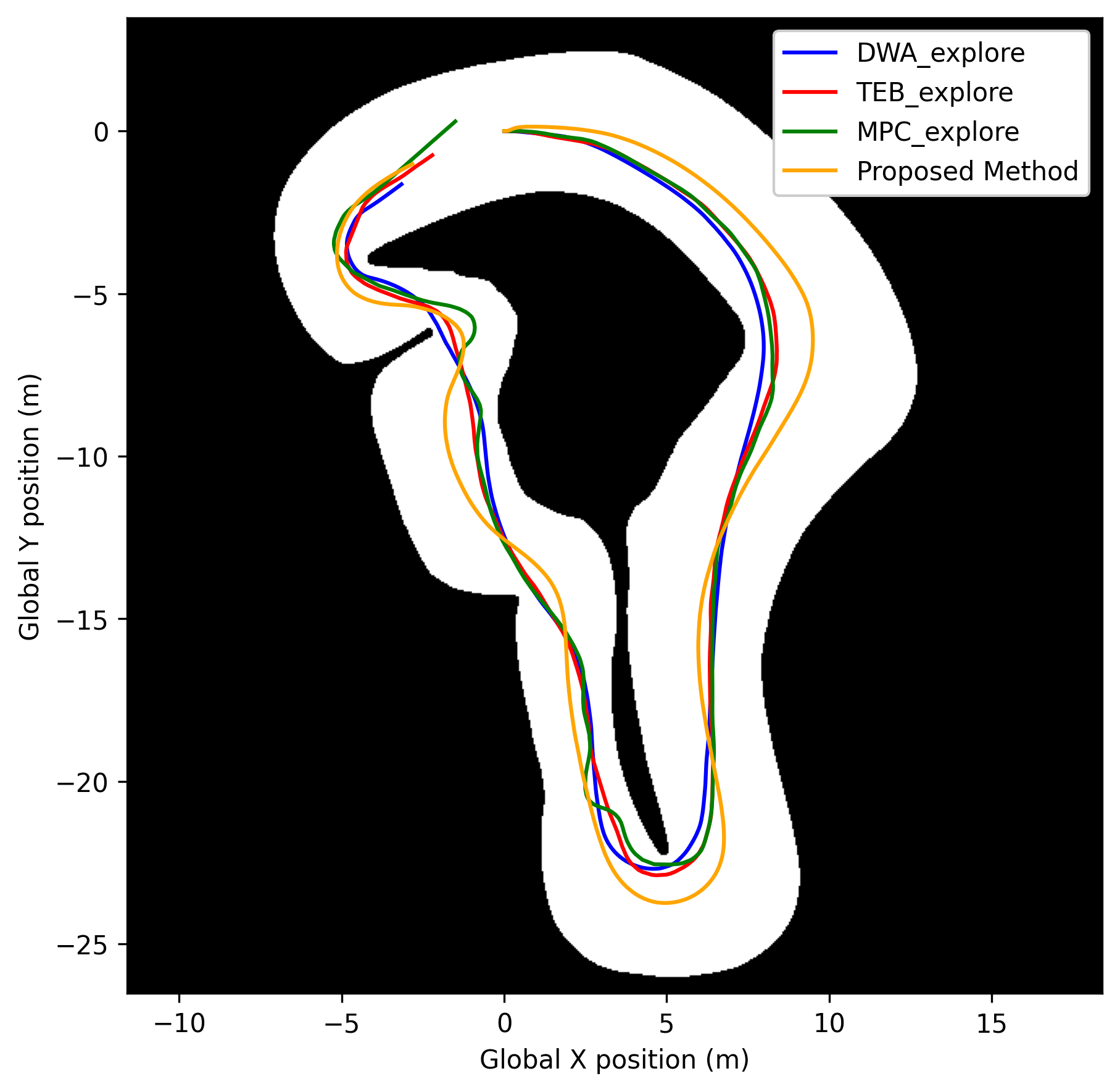}
    \caption{Paths obtained by each local navigation strategy in the simulation. Each path starts at the global origin and proceeds clockwise around the map.}
	\label{PD_sim}
\end{figure}

The path traversed by the proposed approach on a simulated course is compared to those of the DWA, TEB and MPC exploration-based planners in Fig. \ref{PD_sim}. The novel method effectively maintains a path near the center of the track throughout, maximizing obstacle clearance according to its
formulation. Meanwhile, the exploration methods do not attain central paths along the course, instead cutting corners and risking closer obstacle proximity. Of the exploration planners, local planning using TEB is
most effective, as DWA progresses at a lower speed than
desired, while MPC occasionally experiences unexpected oscillations due to poor optimization solutions.

Table \ref{tab:PDsim3} measures navigation performance in simulation according to several key metrics. The sample time is denoted by $\tilde k\in\{0,...,\tilde k_\textup{end}\}$, ranging over the full test, where $d_{\textup{min},\tilde k}$ indicates the minimum obstacle proximity to the current vehicle position at sample $\tilde k$. The minimum obstacle proximity over the full test is represented by $\displaystyle\min_{\tilde k} d_{\textup{min},\tilde k}$, while $\bar{d}_\textup{min}$ denotes the average minimum proximity. For both control inputs, the mean ($\overline{|\delta_{\textup{cmd},\tilde k}|},\bar{v}_{\textup{cmd},\tilde k}$) and variance ($\mathrm{Var}(\delta_{\textup{cmd},\tilde k}),\mathrm{Var}(v_{\textup{cmd},\tilde k})$) are also reported. Notably, the proposed navigation method achieves superior safety through a higher $\displaystyle\min_{\tilde k} d_{\textup{min},\tilde k}$ and especially $\bar{d}_\textup{min}$ compared to other planners while attaining low control effort and a velocity near the desired nominal value.

\begin{table}[ht]
  \caption{Simulation performance of the proposed navigation method and existing local planners}
  \label{tab:PDsim3}
  \centering
  \begin{tabular}{l@{\hskip -1.0pt}c@{\hskip 2pt}c@{\hskip 5pt}c@{\hskip 4pt}c@{\hskip 3pt}c@{\hskip 3pt}c}
    \toprule\\[-0.85em]
    \scriptsize Local Planner & \makecell{\scriptsize$\min_{\tilde k} d_{\textup{min},\tilde k}$\\[0.2em](m)} & \makecell{\scriptsize$\bar{d}_\textup{min}$\\[0.3em](m)} & \makecell{\scriptsize$\overline{|\delta_{\textup{cmd},\tilde k}|}$\\[0.2em](rad)} & \makecell{\scriptsize$\mathrm{Var}(\delta_{\textup{cmd},\tilde k})$\\[0.2em]($\textup{rad}^2$)} & \makecell{\scriptsize$\bar{v}_{\textup{cmd},\tilde k}$\\[0.3em](m/s)} & \makecell{\scriptsize$\mathrm{Var}(v_{\textup{cmd},\tilde k})$\\[0.2em]($\textup{m}^2$/$\textup{s}^2$)}\\
    \midrule\\[-0.85em]
    \scriptsize DWA\_explore      & 0.207 & 1.147 & 0.060 & 0.008 & 0.492 & 0.011 \\
    \scriptsize TEB\_explore     & 0.521 & 1.289 & 0.075 & 0.011 & 1.474 & 0.027  \\
    \scriptsize MPC\_explore   & 0.278 & 1.268 & 0.124 & 0.032 & 1.246 & 0.201  \\
    \scriptsize Proposed Method   & 0.508 & 1.677 & 0.080 & 0.012 & 1.482 & 0.009  \\
    \bottomrule
  \end{tabular}
\end{table}

\subsection{Experimental Results}
\begin{figure}[ht]
    \centering
    \includegraphics[width=0.89\columnwidth]{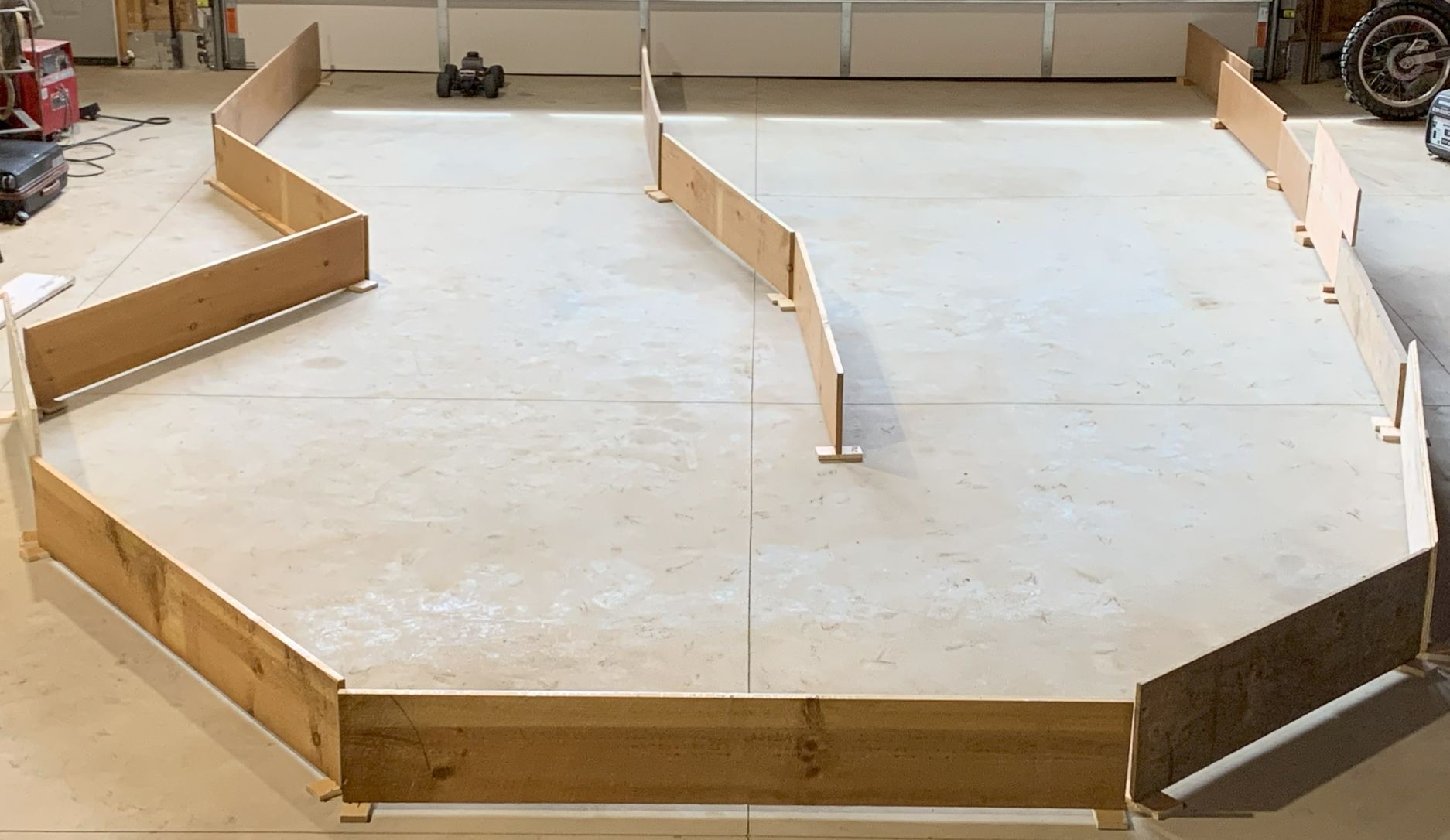}
    \caption{Experiment course layout and $1/10^\textup{th}$ scale RC vehicle.}
	\label{ExperimentMap}
\end{figure}

\begin{figure}[ht]
    \centering
    \includegraphics[width=0.94\columnwidth]{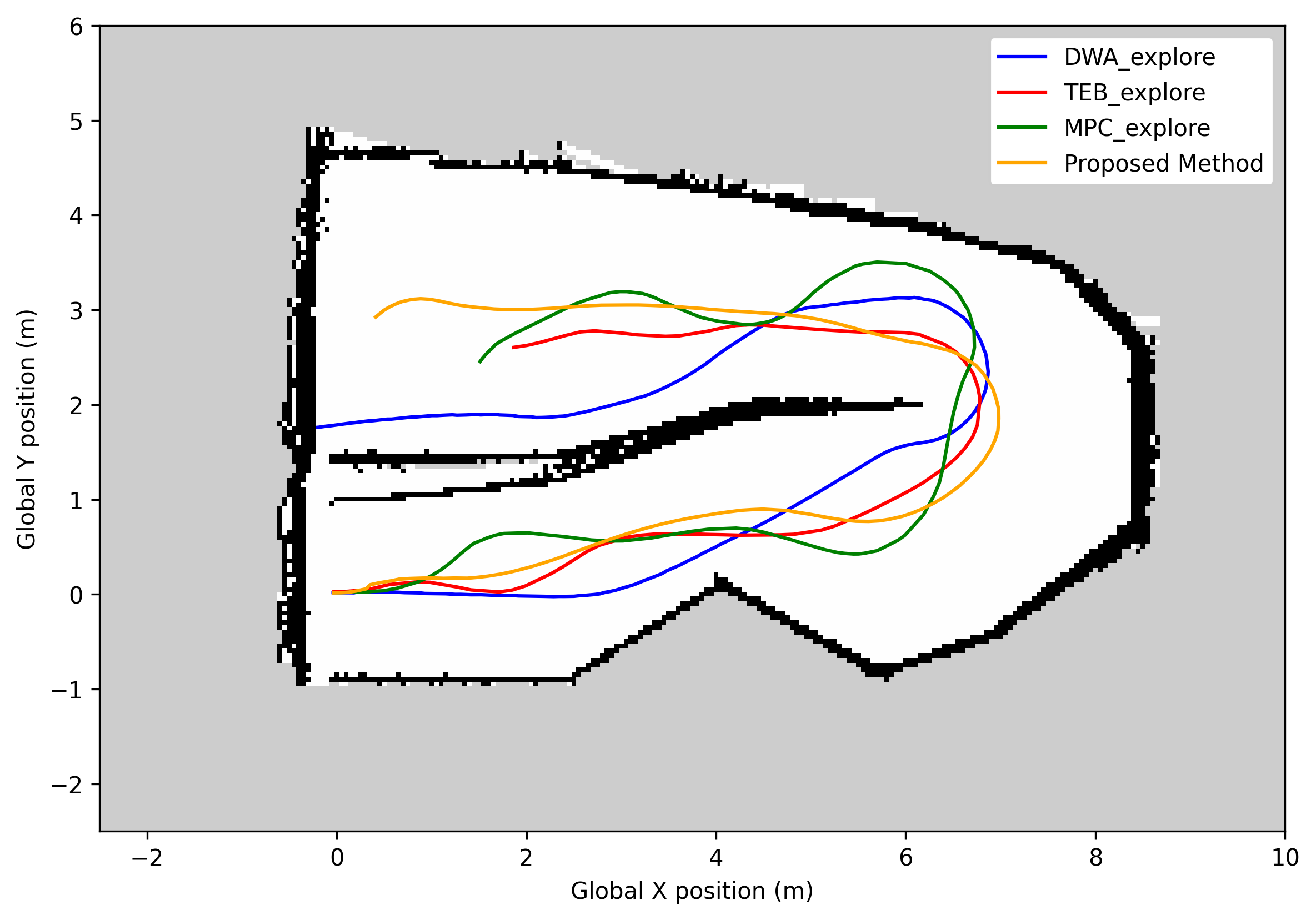}
    \caption{Paths obtained by each local navigation strategy in the experiment. Each path starts at the global origin and navigates the map counterclockwise.}
	\label{PD_exp}
\end{figure}

Moreover, testing is conducted using a $1/10^\textup{th}$ scale RC vehicle on the course layout illustrated in Fig. \ref{ExperimentMap}. Here, local navigation performance is validated in real-world conditions with imperfect sensor capabilities and vehicle dynamics that are not fully captured by the kinematic bicycle model. The same parameter values used in Section \ref{simsec}'s simulation are retained in the experiment. Again, the proposed navigation method produces a smooth path, while more effectively maximizing obstacle clearance compared to the exploration-based planners (Fig. \ref{PD_exp}).

Using the key metrics detailed in Section \ref{simsec}, navigation performance in the experiment is quantified via Table \ref{tab:PDexp3}. Mirroring the simulation results, the proposed method performs best in terms of obstacle proximity metrics and exhibits low control effort. The average velocity attained is close to the desired nominal value, but lower in practice due to initial acceleration, deceleration upon course completion, and the incorporated vehicle slowdown function. Meanwhile, the DWA and MPC exploration-based local planners attain significantly lower average velocities and poorer minimum obstacle clearance measures compared to the best baseline method, TEB. Thus, the proposed navigation scheme is transferable to real-world conditions, where effective performance and safety are maintained.   

\begin{table}[ht]
  \caption{Real-world performance of the proposed navigation method and existing local planners}
  \label{tab:PDexp3}
  \centering
  \begin{tabular}{l@{\hskip -1.0pt}c@{\hskip 2pt}c@{\hskip 5pt}c@{\hskip 4pt}c@{\hskip 3pt}c@{\hskip 3pt}c}
    \toprule\\[-0.85em]
    \scriptsize Local Planner & \makecell{\scriptsize$\min_{\tilde k} d_{\textup{min},\tilde k}$\\[0.2em](m)} & \makecell{\scriptsize$\bar{d}_\textup{min}$\\[0.3em](m)} & \makecell{\scriptsize$\overline{|\delta_{\textup{cmd},\tilde k}|}$\\[0.2em](rad)} & \makecell{\scriptsize$\mathrm{Var}(\delta_{\textup{cmd},\tilde k})$\\[0.2em]($\textup{rad}^2$)} & \makecell{\scriptsize$\bar{v}_{\textup{cmd},\tilde k}$\\[0.3em](m/s)} & \makecell{\scriptsize$\mathrm{Var}(v_{\textup{cmd},\tilde k})$\\[0.2em]($\textup{m}^2$/$\textup{s}^2$)}\\
    \midrule\\[-0.85em]
    \scriptsize DWA\_explore      & 0.385 & 0.654 & 0.118 & 0.028 & 0.491 & 0.012 \\
    \scriptsize TEB\_explore     & 0.570 & 0.843 & 0.157 & 0.032 & 1.356 & 0.139  \\
    \scriptsize MPC\_explore   & 0.313 & 0.831 & 0.253 & 0.073 & 0.858 & 0.149  \\
    \scriptsize Proposed Method   & 0.587 & 0.978 & 0.147 & 0.032 & 1.281 & 0.107  \\
    \bottomrule
  \end{tabular}
\end{table}

Computation times ($t_{\textup{comp},\tilde k}$) for each navigation approach are evaluated in terms of both the mean ($\bar{t}_\textup{comp}$) and maximum ($\displaystyle\max_{\tilde k} t_{\textup{comp},\tilde k}$, abbreviated as $\displaystyle\max t_\textup{comp}$) and summarized in Table \ref{tab:PDcomp3}. For the exploration-based planners, computation occurs across three stages: updating unexplored frontiers and selecting a new local goal, performing global planning toward this goal and finally, executing local planning to follow this path while avoiding collisions. The third stage varies between the different exploration methods. Computation times for the individual stages are represented as $t_{\textup{frn},\tilde k},\,t_{\textup{glb},\tilde k}\ \& \ t_{\textup{loc},\tilde k}$, where their sum equals the total planning computation time $t_{\textup{comp},\tilde k}$.

The proposed navigation method achieves the lowest average and worst-case computation times due to the computational efficiency of convex optimization and feedback-linearizing control. The computation times of all other planners are, at best, on the order of magnitude greater than that of the proposed approach, although each still satisfies the 100-ms control period on average. MPC-based exploration significantly exceeds this control period in the worst case, when an optimization solution cannot be readily obtained and no optimization timeout is enforced to maintain real-time operation.

\begin{table}[ht]
\caption{Real-world planning computation times for the proposed navigation method and existing local planners}
\label{tab:PDcomp3}
\centering
\begin{tabular}{
    l@{\hskip -1.0pt}c@{\hskip 3.2pt}c@{\hskip 3.2pt}c@{\hskip 4.5pt}c@{\hskip 2.4pt}c@{\hskip 2.4pt}c@{\hskip 2.4pt}c@{\hskip 2.4pt}c
}
\toprule\\[-1.2em]
\scriptsize Local Planner &
\makecell{\scriptsize$\bar{t}_\textup{frn}$\\(ms)} &
\makecell{\scriptsize$\bar{t}_\textup{glb}$\\(ms)} &
\makecell{\scriptsize$\bar{t}_\textup{loc}$\\(ms)} &
\makecell{\scriptsize\bm{$\bar{t}_\textbf{comp}$}\\\textbf{(ms)}} &
\makecell{\scriptsize$\max t_\textup{frn}$\\(ms)} &
\makecell{\scriptsize$\max t_\textup{glb}$\\(ms)} &
\makecell{\scriptsize$\max t_\textup{loc}$\\(ms)} &
\makecell{\scriptsize$\bm{\max t_\textbf{comp}}$\\\textbf{(ms)}} \\
\midrule\\[-1.2em] 
{\scriptsize DWA\_explore} & 3.6 & 0.9 & 11.8 & \textbf{16.3} & 7.4 & \phantom{0}1.9 & \phantom{0}27.9 & \phantom{0}\textbf{37.1}  \\
{\scriptsize TEB\_explore} & 1.9 & 0.8 & \phantom{0}3.2 & \phantom{0}\textbf{5.9} & 5.2 & \phantom{0}1.5 & \phantom{00}8.3 & \phantom{0}\textbf{15.0}  \\
{\scriptsize MPC\_explore} & 2.0 & 1.1 & 43.7 & \textbf{46.7} & 3.8 & 10.2 & 484.8 & \textbf{498.8}  \\
{\scriptsize Proposed Method} & \multicolumn{3}{c}{~~~} & \phantom{0}\textbf{0.5} & \multicolumn{3}{c}{~~~} & \phantom{00}\textbf{1.7}  \\
\bottomrule
\end{tabular}
\end{table}

\section{Conclusion}
This paper outlined a novel approach for safe, local navigation in unknown environments for mobile robots with Ackermann steering. A process for finding the safest local heading direction was established before obstacle bounding lines were constructed through several unique obstacle clearance maximization schemes. A feedback-linearizing PD controller then regulated the vehicle position relative to either the left, right or both obstacle bounding lines through the steering angle control input. High collision-risk obstacles directly ahead motivated velocity control for vehicle slowdown, before the steering angle and velocity control inputs were constrained by vehicle dynamics and applied for actuation.

In simulation and experiment, this navigation method produced the safest paths with low control effort and achieved the lowest computation times compared to some existing exploration-based approaches. Therefore, this method serves as a basis for universal safe local navigation in unmapped settings and offers the opportunity for integration in a hierarchical planner for safe global navigation. It should be noted that although the method proposed here was developed for vehicles with Ackermann steering, it could be revised to accommodate mobile robots with different motion kinematics.

\bibliographystyle{IEEEtran}
\bibliography{references}

\end{document}